\documentclass[lettersize,journal]{IEEEtran}
\usepackage{amsmath,amsfonts}
\usepackage{algorithmic}
\usepackage{algorithm}
\usepackage{array}
\usepackage[caption=false,font=normalsize,labelfont=sf,textfont=sf]{subfig}
\usepackage{textcomp}
\usepackage{stfloats}
\usepackage{url}
\usepackage{verbatim}
\usepackage{graphicx}
\usepackage{cite}
\usepackage{tikz}
\usepackage{comment}
\usepackage{mathrsfs}
\usepackage{amsfonts,bm}
\usepackage{times}

\usepackage{soul}
\usepackage{url}
\usepackage[utf8]{inputenc}
\usepackage{amsmath}
\usepackage{booktabs}
\usepackage{color}
\usepackage[normalem]{ulem}
\useunder{\uline}{\ul}{}
\usepackage{bbding}
\usepackage{amsmath}
\usepackage{booktabs}
\usepackage{color}
\usepackage[normalem]{ulem}
\useunder{\uline}{\ul}{}
\newtheorem{theorem}{Theorem}
\newtheorem{definition}{Definition}
\begin{document}

\title{Jacobian Norm for Unsupervised Source-Free Domain Adaptation}

\author{Weikai~Li, Meng Cao and Songcan~Chen% <-this % stops a space
\IEEEcompsocitemizethanks{\IEEEcompsocthanksitem The authors are with College of Computer Science and Technology, Nanjing University of Aeronautics and Astronautics of (NUAA), Nanjing, 211106, China.\protect\\
E-mail: \{leeweikai; alarsh; s.chen\}@nuaa.edu.cn.
\IEEEcompsocthanksitem Corresponding author is Songcan Chen.}% <-this % stops an unwanted space

\thanks{This paper was produced by the IEEE Publication Technology Group. They are in Piscataway, NJ.}% <-this % stops a space
\thanks{Manuscript received April 19, 2021; revised August 16, 2021.}}

\markboth{Journal of \LaTeX\ Class Files,~Vol.~14, No.~8, August~2021}%
{Shell \MakeLowercase{\textit{et al.}}: A Sample Article Using IEEEtran.cls for IEEE Journals}

\IEEEpubid{0000--0000/00\$00.00~\copyright~2021 IEEE}

\maketitle

\begin{abstract}
 Unsupervised Source (data) Free domain adaptation (USFDA) aims to transfer knowledge from a well-trained source model to a related but unlabeled target domain. In such a scenario, all conventional adaptation methods that require source data fail. To combat this challenge, existing USFDAs turn to transfer knowledge by aligning the target feature to the latent distribution hidden in the source model. However, such information is naturally limited. Thus, the alignment in such a scenario is not only difficult but also insufficient, which degrades the target generalization performance. To relieve this dilemma in current USFDAs, we are motivated to explore a new perspective to boost their performance. For this purpose and gaining necessary insight, we look back upon the origin of the domain adaptation and first theoretically derive a new-brand target generalization error bound based on the model smoothness. Then, following the theoretical insight, a general and model-smoothness-guided Jacobian norm (JN) regularizer is designed and imposed on the target domain to mitigate this dilemma. Extensive experiments are conducted to validate its effectiveness. In its implementation, just with a few lines of codes added to the existing USFDAs, we achieve superior results on various benchmark datasets.
\end{abstract}


\section{Introduction}
\IEEEPARstart{D}{eep} neural networks have achieved remarkable success in various multi-media applications, where sufficiently large-scale and well-labeled data are present. However, manually labeling sufficient data is often time-consuming and labor-exhaustive, thus hard to meet the demand of rapid growth of the multi-media steaming or the content sharing applications \cite{yao2019heterogeneous}. To address it, unsupervised domain adaptation (UDA) is becoming an increasingly attractive research topic in the multi-media community \cite{kouw2019review,li2020unsupervised,li2019joint,wang2020prototype,wang2021interbn}. % UDA 成为一种有效的手段
Specifically, by leveraging the discriminative knowledge from readily available and labeled source domains, UDA aims to establish a desired prediction model for the unlabeled target domain, which significantly relieves the burden of annotating magnanimous data. In the past few years, UDA has achieved several impressive results on various multi-media applications such as image recognition \cite{long2018conditional,jing2020adaptively}, semantic segmentation \cite{cheng2021dual}, text classification \cite{guo2020multi}, recommendation \cite{jiang2017deep} and action recognition \cite{luo2020adversarial}.

Currently, a series of theoretical results have been presented to guide the solution for addressing UDA \cite{ben2007analysis,ben2010theory}, which illustrates that the target risk is bounded by the between-domain discrepancy. Motivated by these theoretical results, domain alignment comes to the dominant strategy for solving domain adaptation \cite{kouw2019review}, whose goal is to alleviate the between-domain discrepancy, so that the learned source model can be naturally adapted to the target domain. Specifically, these methods can be categorized into the sample alignment or the feature alignment \cite{kouw2019review}. In particular, the sample alignment focuses on mitigating the domain shifts by the importance-weighting of samples \cite{tsuboi2009direct,sugiyama2008direct,huang2007correcting}. In contrast, the feature alignment attempts to learn a domain invariant feature or representation to alleviate the domain discrepancies by kernel matching \cite{pan2010domain}, adversarial learning \cite{ganin2015unsupervised}, prototype matching \cite{wang2020prototype}, optimal transport \cite{deng2021informative} and image reconstruction \cite{2018Generate}.

 In practice, almost all existing UDA approaches require to access both raw source and target data while learning to adapt. Unfortunately, due to the cost of storage or the protection of private information, in reality, the source data is often not at hand. In light of this, a more realistic scenario, called Unsupervised Source-Free Domain Adaptation (USFDA), is considered, in which only source model can be accessed at adaptation \cite{liang2020we,li2020model,ye2021source}. In such a scenario with no source data, the explicit sample alignment approaches completely fail. \textcolor{black}{ Instead, the existing USFDA approaches attempt to transfer knowledge by \textit{implicitly} aligning the target features with the source by pseudo labeling \cite{liang2020we} or image generation \cite{li2020model}, which are derived from the source model.  However, %the pseudo labelling approach may suffer wrong labelling due to the domain shift. Meanwhile, generating images from the source model is very complicated, even difficult. Digging deeper into these issues, we believe the main reason is that the information from the source model is restricted and limited. This motivates us to mine more information from the target domain itself to boost the performance of the USFDA.
 focusing on implicit alignment alone is often insufficient to address USFDA, since obtaining a desired alignment is usually difficult in the absence of source data, while such an implicit alignment cannot guarantee the success of the adaptation \cite{Siry2021inductive}. To relieve such dilemma, we are motivated to look back upon the origin of USFDA: \textit{ how to guarantee the generalization performance of the learned model on target domain, without accessing to the original source samples?} }
 For this purpose, we aim to explore a new insight to relieve this dilemma in the existing USFDAs. Specifically, %inspired by the benefits of the inductive bias for solving domain adaptation \cite{Siry2021inductive}, we aim to mine the precious knowledge from the target domain by incorporating an appropriate inductive bias. Specifically, we go back to the origin of the domain adaption , which can reduce the target empirical risk and thereby boost the knowledge transfer. To achieve it, 
%挖掘源模型知识，同时
% 因此，模型平滑项可以有效的作为现有基于隐式对齐方法的补充，
%考虑到其简单性，
%仅仅利用这些方式是不足的，
a novel target generalization error bound is derived, which incorporates a between-domain discrepancy term and a new model-smoothness term. In contrast to the existing theoretical works, such a new term is the first time to be considered in adaptation. This term indicates that the model output should be smooth/consistent in the neighborhood of target sample, which instructs us to utilize such neighbor information for guiding the knowledge transfer. It should be noted that the existing USFDAs purely focus on mitigating the between-domain discrepancy, while all of them neglect the neighbor information (i.e., model smoothness). Thus, motivated by the theoretical perspective, we focus on optimizing this term on the target domain to boost the adaptation ability of the existing USFDAs.

Driven by this, a quite simple and model-smoothness-derived Jacobian Norm (JN) regularizer is designed as a plug-in unit to mitigate the dilemma and further boost the performance in the existing USFDAs and used to \textit{implicitly} force the smoothness/consistency of the model output in the neighborhood of target sample. Consequently, it takes advantage of both the target data and the source model, as shown in Figure \ref{fig1}. Due to the convenience of the pseudo-based strategy, we adopt the prevailing pseudo-labelling approach as a baseline to potentially reduce the between-domain discrepancy. In its implementation, we freeze the pre-trained classifier and fine-tune the feature encoding module by minimizing the JN regularized objective to boost its performance. In the end, we conduct abundant experiments on several domain adaptation datasets with different sample sizes. The experimental results demonstrate significant superiority of our model in USFDA. The contributions of this paper are as follows:
\begin{figure*}[t]
  \centering
  \includegraphics[width=\textwidth]{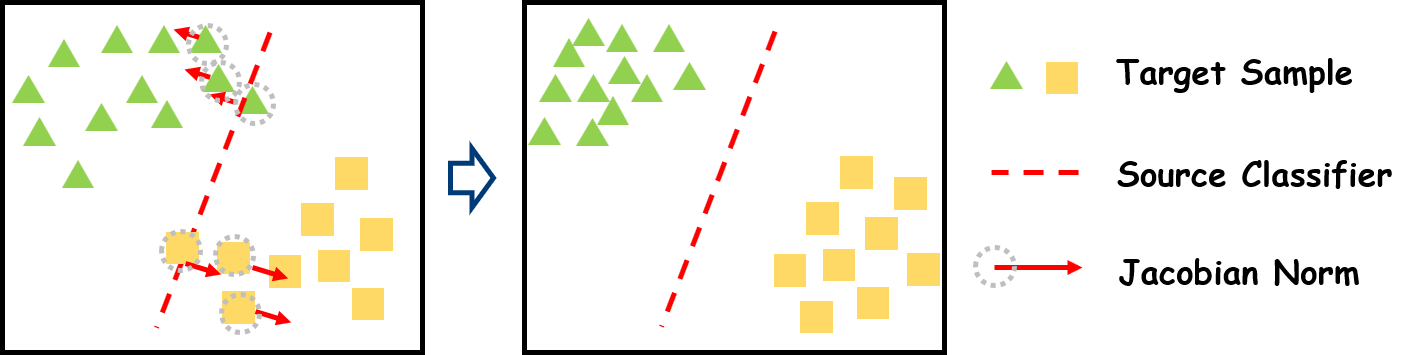}
    \caption{We propose a Jacobian norm regularizer, which effectively utilizes the neighbor information of the target sample to boost the performance of USFDA.}
  \label{fig1}
\end{figure*}

%  简单 只需几行代码
% 容易扩展，
% 有效
 \begin{itemize}
\item We theoretically provide a new-brand target generalization bound based on the Total Variation distance \cite{villani2009optimal} and model smoothness, which provides a novel insight for solving USFDA.
\item We develop a simple yet general JN regularizer to boost the performance of USFDA, which can be easily incorporated into any existing USFDA methods as a plug-in unit with a few lines of code increased.
\item We empirically find that JN regularizer can significantly improve the performance of the existing USFDA method, which achieves competitive results on multiple datasets.
\end{itemize}

The remainder of this paper is organized as follows. In Section \ref{section2}, we briefly overview unsupervised domain adaptation and unsupervised source-free domain adaptation. In Section \ref{section3}, we present the problem definition and derive a new-brand target generalization error bound. In Section \ref{sec4}, We develop the JN regularizer and the entire USFDA model to address USFDA. The experimental results and the post-hoc analysis are reported in Section \ref{sec5}. In the end, we conclude the entire paper with future research directions in Section \ref{sec6}.
% 在没有东西可以用的情况下，我们只能通过目标域自救，来达到自适应的目的

\section{Related Works}

In this section, we present the most related researches on UDA/USFDA and highlight the differences between these methods and ours.
\label{section2}

\subsection{Unsupervised Domain Adaptation}
 Recent practices on UDA usually attempt to minimize the domain discrepancy for knowledge transfer. Following this, multiple domain adaptation techniques have been developed, which can be summarized into the sample alignment  \cite{tsuboi2009direct,sugiyama2008direct,huang2007correcting} and feature alignment \cite{pan2010domain,fernando2013unsupervised,ganin2015unsupervised}. In particular, the sample alignment methods focus on mitigating the between-domain divergence such as the $\mathcal{A}$-distance \cite{huang2007correcting}, Maximum Mean Discrepancy (MMD) \cite{sugiyama2008direct}, or KL-divergence \cite{tsuboi2009direct} through re-weighting the individual samples. In contrast, the feature alignment methods generate the domain-invariant feature through kernel matching \cite{pan2010domain}, adversarial learning \cite{ganin2015unsupervised}, transferrable contrastive learning \cite{chen2021transferrable}, prototype matching \cite{wang2020prototype}, optimal transport \cite{deng2021informative} and image reconstruction \cite{2018Generate}, to reduce the distribution differences across domains, such as MMD \cite{pan2010domain}, central moment discrepancy \cite{zellinger2017central}, Joint MMD \cite{long2017deep}, $\mathcal{A}$-distance \cite{ganin2015unsupervised} and maximum classifier discrepancy \cite{saito2018maximum}, Wasserstein distance \cite{courty2014domain}, etc. 
 % participated as
 
 % 梵高
Compared with the UDA methods which require access to both source and target data, our work does NOT require the source data while learning to adapt. This is more suitable in real-world applications.
 %  具体是怎么实现的，不要说问题。
 \subsection{Unsupervised Source-Free Domain Adaptation}

Different from UDA, USFDA is a more practical scenario in which the source data is inaccessible at adaptation. Existing methods seek to implicitly align the target domain feature to the source domain by heuristically leveraging the information from the source model. The conventional USFDA methods include pseudo labelling (e.g., SHOT \cite{liang2020does} and BAIT \cite{yanga2010casting}), batch normalization (e.g., BN \cite{ishii2021source}) or data generation (e.g., MA \cite{li2020model}). Specifically, the pseudo labelling methods implicitly align representations from the target domains to the source model, the BN method minimizes the discrepancy between domains by BN statistics stored in the source model, and the image generation methods align the target domain to the annotated data generated by source model. 

The existing USFDAs mainly focus on mitigating the between-domain divergence, which often neglect the neighbor information. Instead, the proposed JN regularization term focuses on optimizing the model smoothness to leverage the neighbor information from the target domain itself, which can be used as a plug-in unit to effectively boost performance of the existing USFDAs.
 
%a model is input smooth to input perturbation on training samples (namely, input-robust model), it

\section{Model Smoothness for Target Generalization Error Bound}
The current theoretical works on domain adaptation are typically established on domain discrepancy, which encourages domain alignment in solving domain adaptation \cite{ben2010theory,ben2007analysis}. However, in USFDA, due to the absence of source data, such an alignment is often difficult and insufficient. To boost the adaptation ability of USFDA, we now look back upon the origin of domain adaptation and attempt to induce a novel insights for addressing USFDA.

\label{section3}
\subsection{Preliminaries}
In this paper, we focus on the USFDA task. We use $\mathcal{X} \subset \mathbb{R}^d$ and $\mathcal{Y}\subset \mathbb{R}$ to denote the feature and the label space, respectively. We assume that the feature space $\mathcal{X}$ of source and target domain has a compact support. Thus, there exists a constant $D>0$, such that $\forall {\mathbf{u,v}\in\mathcal{X}},\Vert \mathbf{u-v} \Vert <D$. In particular, we are given $n$ labeled samples $\{x_i^s,y_i^s\}_{i=1}^{n}$ from the source domain $\mathcal{D}_s$ with the distribution $\mathbb{P}$, where $x_i^s\in\mathcal{X}$ and $y_i^s\in \mathcal{Y}$. We also have $m$ unlabeled samples $\{x_i^t\}_{i=1}^{m}$ from $\mathcal{D}_t$ with the distribution $\mathbb{Q}$, where $x_i^t\in\mathcal{X}$.  In the USFDA setting, $\mathbb{P}\neq\mathbb{Q}$ and the source data can be only accessed at the source model training procedure. In particular, we consider the K-way classification task. The goal of the USFDA is to learn a target function $f:\mathcal{X}\rightarrow\mathcal{Y}$ and predict the target label $\{y_i^t\}_{i=1}^{m}$, where $y_i^t\in \mathcal{Y}$, with only target data $\{x_i^t\}_{i=1}^{m}$ and the source function  $f_s:\mathcal{X}\rightarrow\mathcal{Y}$ available in adaptation.

In addition, let $\mathcal{L}(f(\mathbf{x}),y)$ be the continuous and differentiable loss function. Inspired by a current theoretical study \cite{yi2021improved}, we assume that $0\leq \mathcal{L}(f(\mathbf{x}),y) \leq M$ for constant $M$ without loss of generality. Moreover, we denote the  $\mathcal{E}_{\mathbb{P}}\left(f\right) = \mathbb{E}_{\{x,y\}\sim \mathbb{P}}\mathcal{L}(f(x),y)$ as the expected risk of model $f$ over distribution $\mathbb{P}$.
In order to find an alternative to mitigate the dilemma in USFDA. we also need the following definitions of Total Variation distance \cite{villani2009optimal} and model smoothness:

\begin{definition}
\label{def2}
Total Variation distance \textnormal{\cite{villani2009optimal}}: Given two distributions $\mathbb{P}$ and $\mathbb{Q}$. The Total Variation distance $\mathrm{TV} (\mathbb{P},\mathbb{Q})$ between distributions $\mathbb{P}$ and $\mathbb{Q}$ is defined as:
\end{definition}
\begin{equation}
    \mathrm{TV}(\mathbb{P}, \mathbb{Q})=\frac{1}{2} \int_{\mathcal{X}}|d \mathbb{P}(\mathbf{x})-d \mathbb{Q}(\mathbf{x})|.
\end{equation}

\begin{definition}
\label{def3}
Model Smoothness : A model $f$ with parameter $\mathbf{w}$ is $r$-cover with $\epsilon$-smoothness on distribution $\mathbb{P}$, if 
\end{definition}
\begin{equation}
\label{eq2}
    \mathbb{E}_{\mathbb{P}}\left[\sup _{\|\boldsymbol{\delta}\|_{\infty} \leq r}|f( \mathbf{x}+\boldsymbol{\delta})-f(\mathbf{x})|\right] \leq \epsilon.
\end{equation}
where $\|\boldsymbol{\cdot}\|_{\infty}$ is the $l_\infty$-norm.

\noindent
\textcolor{black}{\textbf{Remark:}  Since the source data is unavailable, the current theoretical works based on $\mathcal{H}$-divergence are often limited in the USFDA setting.  Further, such a divergence is not consistent with the current domain alignment work \cite{shui2020beyond} and let alone the existing USFDA works. In contrast, TV distance is more probabilistically interpretable and  consistent with the objective of the existing USFDAs.  In particular,  pseudo-labeling attempts to minimize the KL-divergence between the one-hot encoding and the model output of the target domain\cite{liang2020we}, BN method also minimizes the KL-divergence between domains by utilizing batch normalization (BN) statistics stored in the source model \cite{ishii2021source}. Moreover, the image generation methods align the target domain to the annotated data by GAN, which is proven to be an approximation of TV distance \cite{pan2020loss}. Note that TV distance can be upper bounded in terms of KL-divergence as $TV(\mathbb{P},\mathbb{Q})\leq\sqrt{KL(\mathbb{P},\mathbb{Q})/{2\log e}}$ \cite{nielsen2018guaranteed}. Thus, we conduct the TV distance to analyze the new target generalization error bound.}

\subsection{Generalization Error Bound Via Model Smoothness}
 With the Total Variation distance in Definition \ref{def2} and model smoothness in Definition \ref{def3}, we present our new-brand generalization error bound on target domain in Theorem \ref{the2}. 
 \begin{theorem}
\label{the2}
 Given two distributions $\mathbb{P}$ and $\mathbb{Q}$, if a model $f$ is $2r$-cover with $\epsilon$ smoothness over distributions $\mathbb{P}$ and $\mathbb{Q}$, with probability at least $ 1 - \theta$, we have:
\end{theorem}
\begin{equation}
\label{eq13}
\begin{aligned}
   \mathcal{E}_{\mathbb{Q}}\left(f\right) 
   \leq  \mathcal{E}_{\mathbb{P}}\left(f\right) +2&\epsilon+ 2M\mathrm{TV}(\mathbb{P},\mathbb{Q})\\
    +  M&\sqrt{\frac{\left(2 d\right)^{\frac{2 \epsilon^{2} D}{r^{2}}+1} \log 2+2 \log \left(\frac{1}{\theta}\right)}{m}}\\
      +  M&\sqrt{\frac{\left(2 d\right)^{\frac{2 \epsilon^{2} D}{r^{2}}+1} \log 2+2 \log \left(\frac{1}{\theta}\right)}{n}}\\
       +  M&\sqrt{\frac{\log(1/\theta)}{2m}}.
\end{aligned}
\end{equation} 

The proof of Theorem \ref{the2} is given in the supplementary file. %This result illustrates that the empirical risk of target domain $\mathcal{E}_{\mathbb{Q}}$ is bounded by the source risk $\mathcal{E}_{\mathbb{P}}\left(f\right)$, the model smoothness $\epsilon$ and the domain discrepancy $\mathrm{TV}(\mathbb{P},\mathbb{Q})$. 
In contrast to the existing theoretical works, we utilize the TV-divergence $\mathrm{TV}(\mathbb{P},\mathbb{Q})$ to measure the domain discrepancy, since TV distance is more probabilistically interpretable and consistent with the objective of the existing USFDAs as mentioned. In particular, the model smoothness $\epsilon$ is firstly considered in adaptation, which provides a new perspective to address USFDA. To relieve the dilemma and boost the performance of the USFDA, we intuitively focus more on optimizing the new term $\epsilon$ to leverage the neighbor knowledge from the target domain.

% 包含了两项，域间散度利用最早的工作，平滑利用ＪＮ 
\section{Domain adaptation without Access to Source Data}
\label{sec4}
In this section, following the idea of the theoretical result, we provide a novel learning framework to address USFDA. More precisely, for the model smoothness term, we present a JN regularizer on target domain as a plug-in unit to boost the existing USFDA, which implicitly forces the smoothness of model in the neighborhood of target sample. Subsequently, we adopt pseudo labelling strategy (i.e., SHOT \cite{liang2020we}) as an attempt baseline to handle the domain discrepancy term, due to its simplicity. 

\begin{figure*}[t]
  \centering
  \includegraphics[width=\textwidth]{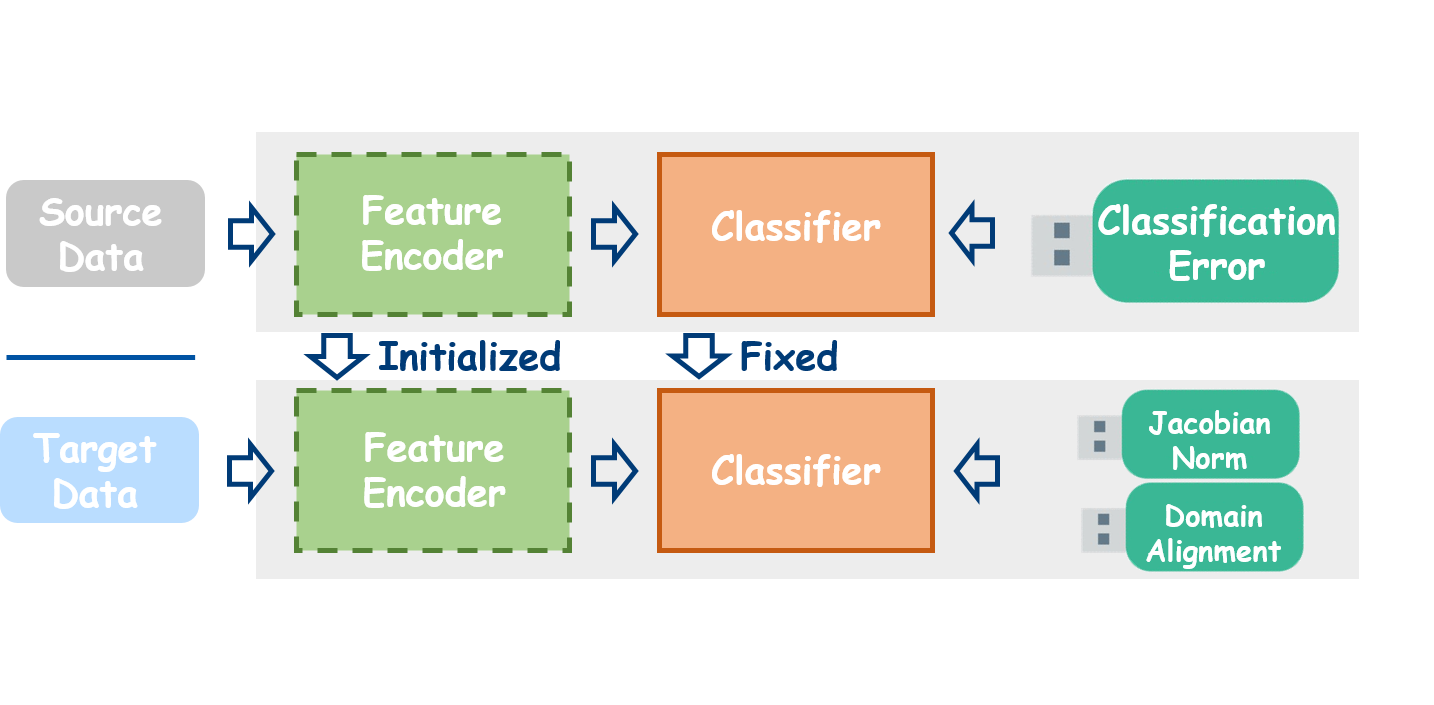}
    \caption{The pipeline of our approach: we freeze (solid line) the source classifier and fine-tune (dash line) the source encoding module. The source data is only available in source model training. The JN regularizer can be easily adopted as a plug-in unit to boost the adaptation performance.}
  \label{fig2}
\end{figure*}

Figure \ref{fig2} shows a pipeline of our approach. Specifically, we first generate the source model by source data. During its development, we keep the classifier frozen and utilize the feature encoding module as initialization for target domain. The feature encoding module is then fine-tuned by our proposed framework. In the following, we elaborate each step of our model.
\subsection{Source Model Generation}
To learn the source model for the subsequent target adaptation, referring to the baseline model \cite{liang2020we}, we adopt the cross-entropy loss based on label smoothing, as it increases the discriminability of the learned source model \cite{muller2019does}
. We have the following objective $\mathcal{L}_{s}$:
\begin{equation}
   \mathcal{L}_{s}=-\mathbb{E}_{\left(x_{s}, y_{s}\right) \in \mathcal{D}_{s}} \sum_{k=1}^{K} q_{k}^{ls} \log \left(\delta_{k}\left(f_{s}\left(x_{s}\right)\right)\right),
\end{equation}
%  SHOT 好的 没有差下来，  SHOT之前差的都搞上去了
where $q_{k}^{ls}=(1-\alpha)q_k+\alpha / K$ is the smoothed label, $q_{k}$ is the one-of $K$ encoding and $\alpha$ is smoothing parameter empirically set to 0.1.  $\delta_{k}\left(\cdot\right)$ denotes the $k$-th element in the soft-max output of a $K$-dimensional vector. 
\subsection{Target Model Fine-tuning}
To make the fixed classifier/model works well in the target domain, we aim to obtain a fine-tuned encoder that could implicitly mitigate the domain discrepancy and optimize the model smoothness on target domain. Therefore, the objective function of our framework is formulated as follows: 
\begin{equation}
\label{eq5}
   \mathcal{L}_{t}=\mathcal{L}_{\mathcal{M}}+\mathcal{L}_{\mathcal{D}},
\end{equation}
where $\mathcal{L}_{\mathcal{M}}$ denotes the model smoothness objective, and  $\mathcal{L}_{\mathcal{D}}$ represents the domain alignment objective. Note that the current USFDAs focus on heuristically modelling the $\mathcal{L}_{\mathcal{D}}$, we aims to formulate  $\mathcal{L}_{\mathcal{M}}$ , which can act as a plug-in unit to any existing USFDAs. These two terms are detailed in the following subsections:
\subsubsection{Jacobian Norm for Model Smoothness}
To formulate $\mathcal{L}_{\mathcal{M}}$, according to the Definition \ref{def3}, we are motivated to control the smoothness/consistence in the neighborhood of the target sample. Along this line, we relax Eq. \ref{eq2} for simplicity, and obtain the following objective:

\begin{equation}
\label{eq6}
    \frac{1}{\sigma^{2}} \sum_{i=1}^{n_{t}} \mathbb{E}_{\zeta}\left(f\left(\mathbf{x}_{i}+\zeta\right)-f\left(\mathbf{x}_{i}\right)\right)^{2},
\end{equation}
where $\zeta \sim N\left(0, \sigma^{2} \mathbf{I}\right)$. By first-order Taylor expansion, and let $\mathbf{J}(\mathbf{x})=\frac{\partial \mathbf{f}}{\partial \mathbf{x}} \in R^{K \times D}$, we have
\begin{equation}
    f(\mathbf{x}+\zeta)=f(\mathbf{x})+\mathbf{J}(\mathbf{x}) \zeta+\boldsymbol{o}(\zeta).
\end{equation}
Omitting the high-order terms, Eq. \ref{eq6} can be reformulated to the following JN regularizer ($\mathcal{L}_{\mathcal{J}}$):
\begin{equation}
\label{eq8}
\begin{aligned}
\mathcal{L}_{\mathcal{J}}&=\frac{1}{\sigma^{2}} \sum_{i=1}^{n_{t}} \mathbb{E}_{\zeta}\left(f \left(\mathbf{x}_{i}\right.\right.\left.+\zeta)-f\left(\mathbf{x}_{i}\right)\right)^{2} \\
&=\frac{1}{\sigma^{2}} \sum_{i=1}^{n_{t}} \mathbb{E}_{\zeta}\left\|\mathbf{J}\left(\mathbf{x}_{i}\right) \zeta\right\|^{2} \\
&=\sum_{i=1}^{n_{t}} \operatorname{tr}\left[\mathbf{J}\left(\mathbf{x}_{i}\right)^{T} \mathbf{J}\left(\mathbf{x}_{i}\right) \frac{1}{\sigma^{2}} \mathbb{E}_{\zeta}\left[\zeta \zeta^{T}\right]\right] \\
&=\sum_{i=1}^{n_{t}} \operatorname{tr}\left[\mathbf{J}\left(\mathbf{x}_{i}\right)^{T} \mathbf{J}\left(\mathbf{x}_{i}\right) \frac{1}{\sigma^{2}} \sigma^{2} \mathbf{I}\right] \\
&=\sum_{i=1}^{n_{t}}\left\|\mathbf{J}\left(\mathbf{x}_{i}\right)\right\|_{F}^{2}.
\end{aligned}
\end{equation}
Consequently, we formulate $\mathcal{L}_{\mathcal{M}}$ to JN regularizer $\mathcal{L}_{\mathcal{J}}$
where $\lambda$ is the balancing hyper-parameters.
\begin{equation}
\label{eqm}
    \mathcal{L}_{\mathcal{M}}=\lambda\mathcal{L}_{\mathcal{J}}=\lambda\sum_{i=1}^{n_{t}}\left\|\mathbf{J}\left(\mathbf{x}_{i}\right)\right\|_{F}^{2}.
\end{equation}
%It is worth noting that the proposed $\mathcal{L}_{\mathcal{J}}$ is only a simple attempt to formulate $\mathcal{L}_{\mathcal{M}}$. The regularizer is flexible enough to be embedded into any other existing USFDAs.  

\subsubsection{Nearest Centroid classifier for Obtaining Pseudo Label}
To alleviate the harmful effects caused by the inaccurate network outputs, we further apply pseudo-labeling for each unlabeled data to better supervise the target data encoding training. Inspired by the DeepCluster \cite{caron2018deep}, we first attain the centroid for each class in the target domain as follows:

\begin{equation}
c_{k}^{(0)}=\frac{\sum_{x_{t} \in \mathcal{X}_{t}} \delta\left(\hat{f}_{t}^{(k)}(x)\right) \hat{g}_{t}(x)}{\sum_{x_{t} \in \mathcal{X}_{t}} \delta\left(\hat{f}_{t}^{(k)}(x)\right)},
\end{equation}
where $\hat{f}_t$ denotes the previously learned target model and $\hat{g}_{t}(x)$ is the target encoder.  Then, we obtain the pseudo labels via the nearest centroid classifier
\begin{equation}
\hat{y}_{t}=\arg \min _{k} D_{f}\left(\hat{g}_{t}\left(x_{t}\right), c_{k}^{(0)}\right),
\end{equation}
where $D_f(\cdot, \cdot)$ measures the cosine distance. In the end, the target centroids is computed via the new pseudo labels:
\begin{equation}
c_{k}^{(1)}=\frac{\sum_{x_{t} \in \mathcal{X}_{t}} \mathbb{I}\left(\hat{y}_{t}=k\right) \hat{g}_{t}(x)}{\sum_{x_{t} \in \mathcal{X}_{t}} \mathbb{I}\left(\hat{y}_{t}=k\right)}, 
\end{equation}
where $\mathbb{I}$ is the index function. The final pseudo label is obtained as followings:

\begin{equation}
    \hat{y}_{t}=\arg \min _{k} D_{f}\left(\hat{g}_{t}\left(x_{t}\right), c_{k}^{(1)}\right),
\end{equation}
\subsubsection{Pseudo Labeling for Implicit Alignment}
To formulate the $\mathcal{L}_{\mathcal{D}}$, referring to the baseline \cite{liang2020we}, we adopt both the information maximization loss $\mathcal{L}_{IM}$ and self-supervised pseudo labelling loss $\mathcal{L}_{SSL}$  toward implicit alignment. Their formulations are as follows, respectively:

\begin{equation}
\label{eq9}
    \mathcal{L}_{IM}=-H\left(\frac{1}{K} \sum_{i}^{K} f\left(x_{i}\right)\right)+\frac{1}{K} \sum_{i}^{K} H\left(f\left(x_{i}\right)\right),
\end{equation}
\begin{equation}
\label{eq10}
    \mathcal{L}_{SSL}=-\mathbb{E}_{\left(x_{t}\right) \in \mathcal{D}_{t}} \sum_{k=1}^{K} \hat{q}_{k} \log \left(\delta_{k}\left(f\left(x_{t}\right)\right)\right),
\end{equation}
where $H(\cdot)$ is the entropy function. $\hat{q}_{k}$ is the one-of-$K$ encoding of the target pseudo labels, which is generated by the nearest centroid classifier. For more details, please refer to \cite{liang2020we}. In this way, we define $\mathcal{L}_{\mathcal{D}}$ as the composition of $\mathcal{L}_{IM}$ and $\mathcal{L}_{SSL}$:

\begin{equation}
\label{eqd}
    \mathcal{L}_{\mathcal{D}} = \beta\mathcal{L}_{IM}+\gamma\mathcal{L}_{SSL},
\end{equation}
%\subsubsection{Overall Objective Function}

%Substituting with Eqs. \ref{eqm} and \ref{eqd} in Eq. \ref{eq5}, the overall objective of our model is formulated as:
%\begin{equation}
%\label{eqo}
%    \mathcal{L}_t=\beta\mathcal{L}_{IM}+\gamma\mathcal{L}_{SSL}+\lambda\mathcal{L}_{\mathcal{J}},
%\end{equation}
where $\beta$ and $\gamma$  are the balancing hyper-parameters and we empirically set $\beta=1$ and $\gamma=1$ in our following experiments. 

Last but not least importantly, we need to state that although our model is based on SHOT, for other USFDA methods, our framework can be easily applied by reformulating $\mathcal{L}_{\mathcal{D}}$. Therefore, the JN regularizer is flexible enough to be embedded into any other existing USFDAs.%the proposed JN regularier has a wider applicability.
\begin{table}[t!]
\centering
\caption{Statistics of the benchmark datasets}
\begin{tabular}{cccc}
\toprule
Dataset   & \#Sample & \#Class & \#Domain \\ \midrule
Office-31  & 4652   & 31   & A, W, D    \\
Office-Home & 15500  & 65   & Ar, Cl, Pr, Rw    \\ 
VisDa-C & 207000   & 12   & Real, Synthesis    \\ 
\bottomrule
\end{tabular}
\label{table1}
\end{table}

\begin{table*}[ht!]
\centering
\caption{Classification accuracies (\%) on small-sized Office-31 dataset (ResNet-50).}
\label{tab2}

\begin{tabular}{cccccc|cccccc}
\toprule
    & DAN  & DANN & CDAN & BSP  & SO   & MA   & BAIT & BN  & SHOT  & JN(o) & Our Model \\\midrule
A$\rightarrow$D & 78.6 & 79.7 & 92.9 & 93.0 & 80.1  & {\ul 92.7} & 92.0 & 89.0 & 92.1 & 88.8  & \textbf{95.9}     \\
A$\rightarrow$W & 80.5 & 82.0 & 94.1 & 93.3 & 76.6 & {\ul 93.7} & \textbf{94.6} & 91.7 & 89.6 & 88.9  & 92.5     \\
D$\rightarrow$A & 63.6 & 68.3 & 71.0 & 74.6 & 60.5 & 75.3 & 74.6 & \textbf{78.5} & 74.1 & 72.7  & {\ul 75.4}     \\
D$\rightarrow$W & 97.1 & 96.9 & 98.6 & 98.2 & 95.5 & {\ul 98.5} & 98.1 & \textbf{98.9} & {\ul 98.5} & 98.4  & {\ul 98.5}     \\
W$\rightarrow$A & 62.8 & 67.4 & 69.3 & 72.6 & 63.4 & \textbf{77.8} & 73.9 & 76.6 & 73.3 & 73.4  & {\ul 77.4}     \\
W$\rightarrow$D & 99.6 & 99.1 & 100  & 100  & 98.5 & 98.5 & \textbf{100}  & 99.8 &{\ul 99.9}  & {\ul 99.9}  & {\ul 99.9}     \\
AVE & 80.4 & 82.2 & 87.7 & 88.5 & 79.1& {\ul 89.6} & 89.1 & 89.0 & 87.9  & 87.0  & \textbf{89.9}   \\
\bottomrule

\multicolumn{12}{p{13cm}}{\footnotesize{Due to the simplicity and the availability of source code, we only conduct SHOT as the baseline method. Specifically, our model is actually a JN regularized objective of SHOT, according to Eq.\ref{eqo}. The best accuracy is presented in bold and the second best is underlined, similarly hereinafter.
}}

%\begin{tabular}{cccccccc}
%%\toprule
%Methods  & A$\rightarrow$D &A$\rightarrow$W&D$\rightarrow$A&D$\rightarrow$W& W$\rightarrow$A &W$\rightarrow$D & AVE  \\\midrule
%ResNet50 & 68.9 & 68.4 & 62.5 & 96.7 & 60.7 & 99.3 & 76.1 \\
%DAN            & 78.6 & 80.5 & 63.6 & 97.1 & 62.8 & 99.6 & 80.4 \\
%DANN           & 79.7 & 82.0 & 68.2 & 96.9 & 67.4 & 99.1 & 82.2 \\
%CDAN           & 92.9 & 94.1 & 71.0 & 98.6 & 69.3 & 100  & 87.7 \\
%BSP            & 93.0 & 93.3 & 73.6 & 98.2 & 72.6 & 100  & 88.5 \\
%SO             & 80.1 & 76.6 & 60.5 & 95.5 & 63.4 & 98.5 & 79.1 \\\midrule
%SHOT           & 92.1 & 89.6 & 74.1 & 98.5 & 73.3 & 99.9 & 87.9 \\
%MA             & 92.7 & 93.7 & 75.3 & 98.5 & 77.8 & 98.5 & 89.6 \\
%BAIT           & 92.0 & 94.6 & 74.6 & 98.1 & 73.9 & 100.0 & 89.1 \\
%BN             & 89.0 & 91.7 & 78.5 & 98.9 & 76.6 & 99.8 & 89.0 \\
%JN(o)          & 88.8 & 88.9 & 72.7 & 98.4 & 73.4& 99.9 & 87.0 \\
%JN(ours)       & 95.9 & 92.5 & 75.4 & 98.5 & 77.4 & 99.9 & 89.9\\\bottomrule
\end{tabular}

%\begin{flushleft}
%\footnotesize{Our method is actually a JN regularized objective for SHOT. Here, we only focus on SHOT due to its simplicity and the avialble of source code.}
%\end{flushleft}

\end{table*}

\begin{table*}[ht!]
\centering
\caption{Classification accuracies (\%) on medium-sized Office-Home dataset (ResNet-50).}
\label{tab3}

\begin{tabular}{cccccc|cccc}
\toprule
    & DAN  & DANN & CDAN & BSP  & SO & BAIT   & SHOT & JN(o) & Our Model \\\midrule
Ar$\rightarrow$Cl & 43.6 & 45.6 & 50.7 & 52.0 & 44.9 & \textbf{57.4} & 55.9 & 54.8  & {\ul 56.4}     \\
Ar$\rightarrow$Pr & 57.0 & 59.3 & 70.6 & 68.6 & 66.5 & 77.5& {\ul 77.8} & 76.6  & \textbf{78.6}     \\
Ar$\rightarrow$Re & 67.9 & 70.1 & 76.0 & 76.1 & 74.3 & \textbf{82.4} & 80.7 & {\ul 82.0}  & {\ul 82.0}     \\
Cl$\rightarrow$Ar & 45.8 & 47.0 & 57.6 & 58.0 & 52.9 & 68.0 & 66.9 & \textbf{68.4}  & {\ul 68.3}     \\
Cl$\rightarrow$Pr & 56.5 & 58.5 & 70.0 & 70.3 & 62.8 & {\ul 78.2} & 77.3 & 73.9  & \textbf{80.1}     \\
Cl$\rightarrow$Re & 60.4 & 60.9 & 70.0 & 70.2 & 65.1 & {\ul 78.1} & 77.1 & 76.3  & \textbf{79.4}     \\
Pr$\rightarrow$Ar & 44.0 & 46.1 & 57.4 & 58.6 & 52.9 & 67.4 & 66.4 & {\ul 66.8}  & \textbf{68.8}     \\
Pr$\rightarrow$Cl & 43.6 & 43.7 & 50.9 & 50.3 & 41.5 & \textbf{55.5} & 53.6 & 54.6  & {\ul 55.1}     \\
Pr$\rightarrow$Re & 67.7 & 68.5 & 77.3 & 77.6 & 73.4 & {\ul 81.7} & 81.5 & 81.3  & \textbf{82.2}     \\
Re$\rightarrow$Ar & 63.1 & 63.2 & 70.9 & 72.2 & 65.4 & \textbf{76.3} & 72.1 & 74.2  & {\ul 75.3}     \\
Re$\rightarrow$Cl & 51.5 & 51.8 & 56.7 & 59.3 & 45.4 & {\ul 57.1} & 57.2 & 58.4  & \textbf{58.8}     \\
Re$\rightarrow$Pr & 74.3 & 76.8 & 81.6 & 81.9 & 77.9 & {\ul 84.3} & 83.0 & 83.3  & \textbf{84.6}     \\
AVE & 56.3 & 57.6 & 65.8 & 66.3 & 60.2 & 70.8 & {\ul 71.5} & 70.6  & \textbf{74.5}    \\

%\begin{tabular}{cccccccccccccc}
%\toprule
%Methods  &Ar$\rightarrow$Cl&Ar$\rightarrow$Pr&Ar$\rightarrow$Re&Cl$\rightarrow$Ar&Cl$\rightarrow$Pr&Cl$\rightarrow$Re &Pr$\rightarrow$Ar&Pr$\rightarrow$Cl&Pr$\rightarrow$Re&Re$\rightarrow$Ar &Re$\rightarrow$Cl&Re$\rightarrow$Pr & AVE  \\\midrule
%ResNet50 & 34.9 & 50.0 & 58.0 & 37.4 & 41.9 & 46.2 & 38.5 & 31.2 & 60.4 & 53.9 & 41.2 & 59.9 & 46.1 \\
%DAN      & 43.6 & 57.0 & 67.9 & 45.8 & 56.5 & 60.4 & 44.0 & 43.6 & 67.7 & 63.1 & 51.5 & 74.3 & 56.3 \\
%DANN     & 45.6 & 59.3 & 70.1 & 47.0 & 58.5 & 60.9 & 46.1 & 43.7 & 68.5 & 63.2 & 51.8 & 76.8 & 57.6 \\
%CDAN     & 50.7 & 70.6 & 76.0 & 57.6 & 70.0 & 70.0 & 57.4 & 50.9 & 77.3 & 70.9 & 56.7 & 81.6 & 65.8 \\
%BSP      & 52.0 & 68.6 & 76.1 & 58.0 & 70.3 & 70.2 & 58.6 & 50.2 & 77.6 & 72.2 & 59.3 & 81.9 & 66.3 \\
%SO       & 44.9 & 66.5 & 74.3 & 52.9 & 62.8 & 65.1 & 52.9 & 41.5 & 73.4 & 65.4 & 45.4 & 77.9 & 60.2 \\\midrule
%SHOT     & 55.9 & 77.8 & 80.7 & 66.9 & 77.3 & 77.1 & 66.4 & 53.6 & 81.5 & 72.1 & 57.2 & 83.0 & 70.8 \\
%BAIT     & 57.4 & 77.5 & 82.4 & 68.0 & 78.2 & 78.1 & 67.4 & 55.5 & 81.7 & 76.3 & 57.1 & 84.3 & 71.5 \\
%JN(o)     & 54.8 & 76.6 & 82.0 & 68.4 & 73.9 & 76.3 & 66.8 & 54.6 & 81.3 & 74.2 & 58.4 & 83.3 & 70.6 \\
%JN(ours)       & 56.4 & 78.6 & 82.0 & 68.3 & 80.1 & 79.4 & 68.8 & 55.1 & 82.2 & 75.3 & 58.8 & 84.6 & 74.5\\
\bottomrule
\end{tabular}
\end{table*}
%JN 帶來的增益甚至超過了源於數據

\begin{table*}[ht!]
\centering
\caption{Classification accuracies (\%) on large-sized VisDA-C dataset (ResNet-101).}
\label{tab4}
\begin{tabular}{cccccc|ccccc}
\toprule
       & DAN  & DANN & CDAN & BSP  & SO   & MA   & BAIT & SHOT & JN(o) & Our Model \\\midrule
plane  & 87.1 & 81.9 & 85.2 & 92.4 & 76.1 & {\ul 94.8} & 93.7 & 94.6 & 94.6  & \textbf{95.1}     \\
bcycl  & 63.0 & 77.7 & 66.9 & 61.0 & 21.9 & 73.4 & 83.2 & {\ul 83.1} & 83.4  & \textbf{86.6}     \\
bus    & 76.5 & 82.8 & 83.0 & 81.0 & 51.1 & 68.8 & \textbf{84.5} & 73.3 & {\ul 80.3}  & 78.3     \\
car    & 42.0 & 44.3 & 50.8 & 57.5 & 70.3 & \textbf{74.8} & {\ul 65.0} & 54.3 & 56.8  & 62.1     \\
horse  & 90.3 & 81.2 & 84.2 & 89.0 & 64.6 & {\ul 93.1} & 92.9 & 90.2 & 91.4  & \textbf{94.5}     \\
knife  & 42.9 & 29.5 & 74.9 & 80.6 & 16.0  & {\ul 95.4} & {\ul 95.4}& 67.1 & 92.5  & \textbf{96.4}     \\
mcycl  & 85.9 & 65.1 & 88.1 & 90.1 & 81.2 & \textbf{88.5} & {\ul 88.1}& 78.8  & 84.2  & 84.6     \\
person & 53.1 & 28.6 & 74.5 & 77.0 & 18.5  & \textbf{84.7} & 80.8& 76.3 & 78.3  & {\ul 81.0}     \\
plant  & 49.7 & 51.9 & 83.4 & 84.2 & 69.3  & 89.1 & {\ul 90.0}& 89.6 & 87.6  & \textbf{90.2}     \\
sktbrd & 36.3 & 54.6 & 76.0 & 77.9 & 28.4  & 84.7 & {\ul 89.0}& 87.2 & 88.9  & \textbf{89.5}     \\
train  & 85.8 & 82.8 & 81.9 & 82.1 & 84.3  & 83.6 & 84.0& \textbf{87.3} & \textbf{87.3}  & {\ul 84.7}     \\
truck  & 20.7 & 20.7 & 38.0 & 38.4 & 6.0   & 48.1 & 45.3 & 50.3& {\ul56.2}  & \textbf{58.6}     \\
AVE    & 61.1 & 57.4 & 73.9 & 75.9 & 49.0  & 81.6 & {\ul82.7}& 77.7 & 81.8  & \textbf{83.5}     \\
%\begin{tabular}{cccccccccccccc}
%\toprule
%Methods   & plane & bcycl & bus  & car  & horse & knife & mcycl & person & plant & sktbrd & train & truck & AVE  \\\midrule
%ResNet101 & 55.1  & 53.3  & 61.9 & 59.1 & 80.6  & 17.9  & 79.7  & 31.2   & 81.0  & 26.5   & 73.5  & 8.5   & 52.4 \\
%DAN       & 87.1  & 63.0  & 76.5 & 42.0 & 90.3  & 42.9  & 85.9  & 53.1   & 49.7  & 36.3   & 85.8  & 20.7  & 61.1 \\
%DANN      & 81.9  & 77.7  & 82.8 & 44.3 & 81.2  & 29.5  & 65.1  & 28.6   & 51.9  & 54.6   & 82.8  & 20.7  & 57.4 \\
%CDAN      & 85.2  & 66.9  & 83.0 & 50.8 & 84.2  & 4.9   & 88.1  & 74.5   & 83.4  & 76.0   & 81.9  & 38.0  & 73.9 \\
%BSP       & 92.4  & 61.0  & 81.0 & 57.5 & 89.0  & 80.6  & 90.1  & 77.0   & 84.2  & 77.9.  & 82.1  & 38.4  & 75.9 \\
%SO        & 76.1  & 21.9  & 51.1 & 70.3 & 64.6  & 16.0  & 81.2  & 18.5   & 69.3  & 28.4   & 84.3  & 6.0   & 49.0 \\\midrule
%SHOT      & 94.6  & 83.1  & 73.3 & 54.3 & 90.2  & 67.1  & 78.8  & 76.3   & 89.6  & 87.2   & 87.3  & 50.3  & 77.7 \\
%MA        & 94.8  & 73.4  & 68.8 & 74.8 & 93.1  & 95.4  & 88.5  & 84.7   & 89.1  & 84.7   & 83.5  & 48.1  & 81.6 \\
%BAIT      & 93.7  & 83.2  & 84.5 & 65.0 & 92.9  & 95.4  & 88.1  & 80.8   & 90.0  & 89.0   & 84.0  & 45.3  & 82.7 \\
%JN(o)      & 94.6  & 83.4  & 80.3 & 56.8 & 91.4  & 92.5  & 84.2  & 78.3   & 87.6  & 88.9   & 87.3  & 56.2  & 81.8 \\
%JN        & 95.1  & 86.6  & 78.3 & 62.1 & 94.5  & 96.4  & 84.6  & 81.0   & 90.2  & 89.5   & 84.7  & 58.6  & 83.5\\
\bottomrule
\end{tabular}
\end{table*}
\section{Experiments}
\label{sec5}
In this section, we present the experimental results on multiple domain adaptation benchmarks to demonstrate the effectiveness of our model. 
\subsection{Benchmark Datasets}
To evaluate the performance of our model, we conduct abundant experiments over the most widely-used benchmark datasets \cite{li2019joint} with different sample size including \emph{Office-31} (small-size), \emph{Office-Home} (medium size) and \emph{VisDA-C} (large size). Table \ref{table1} lists the statistics of these datasets. %The details of threse 3 datasets are given below:
\\
\noindent
\textbf{Office-31} (\cite{saenko2010adapting} is a small-size benchmark, which contains 4652 images with 31 categories in three visual domains Amazon(A), DSLR(D), Webcam(W).\\
\noindent
\textbf{Office-Home} \cite{venkateswara2017deep} is a medium-size benchmark, which contains 15588 images of 65 categories from 4 domains: Artistic images (Ar), Clipart images (Cl), Product images (Pr), and Real-world images (Rw).\\
\noindent
\textbf{VisDA-C} (\cite{peng2019domain} is a large-scale benchmark, which contains 207000 images of 12 categories from synthesis and real domains. The source domain contains 152000 of synthetic images, while the target domain has 55000 real object images sampled from Microsoft COCO.
%JN 帶來的增益甚至超過了源於數據

\subsection{Comparison Methods}
To verify the effectiveness of the our work, we compare it respectively with several UDA and USFDA SOTAs. The UDA methods include Deep Adaptation Network (DAN) \cite{long2015learning}, Domain Adversarial Neural Networks (DANN) \cite{ganin2015unsupervised}  Conditional Adversarial Networks (CDAN) \cite{long2018conditional} and Batch Spectral Penalization (BSP) \cite{chen2019transferability}. The USFDA methods contain Source HypOthesis Transfer (SHOT) \cite{liang2020we}, Model Adaptation (MA) \cite{li2020model}, BAIT \cite{yanga2010casting} and Batch Normalization (BN) \cite{ishii2021source}. Moreover, source model only (SO) denotes using the entire source model for target label prediction. JN only (JN(o)) represents using JN regularizer only to fine-tune the feature encoder. \textit{It should be noted that SHOT is the baseline of our model by setting $\lambda=0$. } To evaluate their performance, we follow the widely used \textbf{accuracy} as a measurement. The results of comparison methods are directly obtained from the published papers, since we follow the same setting.  

\subsection{Implementation Details}
\subsubsection{Network architecture}
\textcolor{black}{
Following the current UDA/USFDA works \cite{liang2020we,long2017deep,ganin2015unsupervised}, we employ the pre-trained ResNet-50 or ResNet-101 \cite{he2016deep} models as the backbone module. Specifically, we replace the original FC layer with a bottleneck layer (256 units) and a task-specific FC classifier layer. After that, a BN layer is put inside the bottleneck layer. Moreover, a weight normalization layer is utilized in the last FC layer.
More specifically, for each task, referring to the existing works \cite{long2018conditional,chen2019progressive}, we employ the pre-trained ResNet-50 (\emph{Office-31} and \emph{Office-Home}) or ResNet-101 (\emph{VisDA-C}) \cite{he2016deep} models as the backbone module.} %Following \cite{ganin2015unsupervised}, we replace the original FC layer with a bottleneck layer (256 units) and a task-specific FC classifier layer. 
\subsubsection{Parameter Settings}
\textcolor{black}{
To fine-tune the adaptive model, we adopt the mini-batch SGD with momentum 0.9 and set the batch size as 64. For \emph{Office-31} and \emph{Office-Home}, we empirically set the learning rate as 0.01 and $\lambda=0.2$. Since \emph{VisDA-C} can easily converge, we utilize a smaller learning rate 0.001 and a bigger $\lambda=0.8$. For learning in the target domain, we update the pseudo-labels epoch by epoch. The whole network is trained by the back propagation, while the newly added layers (e.g., task-specific FC classifier layer) are trained with learning rate 10 times of that of the pre-trained layers.
}
\subsection{Experimental Results}

\textcolor{black}{
The experimental results of \emph{Office-31}, \emph{Office-Home}, and \emph{VisDA-C} are reported in Tables \ref{tab2}, \ref{tab3}, and \ref{tab4}, respectively. From these results, we can make several observations as follows.}

\textcolor{black}{
Firstly, by adding a simple JN regularization term, our model obtains the best mean accuracy on \emph{Office-31} and  \emph{Office-Home}, and the best per-class accuracy on \emph{Visda-Home}. Compared with the USFDA SOTAs \cite{li2020model,yanga2010casting,ishii2021source}, \textbf{we achieve the best/second-best results on 5 out of 6 individual tasks at \emph{Office-31} dataset and the best/second best on all 12 tasks at \emph{Office-Home} dataset, respectively. For large-scale synthesis-to-real \emph{VisDA-C} dataset, we achieve the best/second-best class accuracy among 9 out of 12 classes.} These results obtained from a wide range of datasets with different sample sizes demonstrate that our model is capable of reducing the target risk while solving USFDA to great extent.}

\textcolor{black}{
Secondly, compared to the conventional UDA works, we also achieve the competitive results even with no direct access to the source domain data. Specifically, \textbf{our model achieves better gains with 1.4$\%$, 8.2$\%$ and 7.6$\%$ in performance than the UDA SOTAs} on the \emph{Office-31}, \emph{Office-Home} and \emph{VisDA-C} datasets, respectively. This implying the superior of our model even without source data.}

\textcolor{black}{
Thirdly, compared to the baseline model (i.e, SHOT) which only achieves the second best results on two tasks at \emph{Office-31} dataset and one task at \emph{Office-Home}, \textbf{the designed JN regularizer provides gains on almost all tasks (e.g., $\sim$ 4\% on A$\rightarrow$D and W$\rightarrow$A tasks)}. Moreover, the performance of our model only degrades in class 'train' on \emph{VisDA-C}, and the main reason may be that the background of this class is too complex. }
\begin{table}[t!]
\centering
\caption{Ablation Study: Components of Each Method}
\begin{tabular}{cccc}
\toprule
Dataset   & SHOT & JN(o) & Our Model \\ \midrule
Model Smoothness  &\XSolidBrush& \Checkmark &\Checkmark \\
Implicit Alignment & \Checkmark &\XSolidBrush & \Checkmark \\ 
\bottomrule
\end{tabular}
\label{table5}
\end{table}
\subsection{Evaluation of Each Component}
\textcolor{black}{
When solving the USFDA, our model involves two components: (1) Jacobian norm for model smoothness and (2)pseudo labeling for implicit alignment. To verify the performance of different components, we empirically select different components as shown in Table \ref{table5}. Specifically, SHOT only considers the implicit alignment, JN(o) only considers the model smoothness while our model contains both two terms.}

\textcolor{black}{
As expected, compared to SHOT (i.e., $\mathcal{L}_{\mathcal{D}}$ only), \textbf{the proposed JN regularizer provides a significant performance gain (i.e., 2\% over \emph{Office-31}, 3.7\% over \emph{Office-Home} and 5.8\% over \emph{VisDA-C})}, which illustrates that implicit alignment is not sufficient to address USFDA and the proposed JN regularizer can effectively boost the performance of USFDA. Moreover,the results of JN alone (i.e., $\mathcal{L}_{\mathcal{J}}$ only) also achieves comparable results on USFDA with the accuracy of 87.0\%, 70.6\% and 81.8\% on the \emph{Office-31}, \emph{Office-Home} and \emph{VisDA-C}, respectively. }

\textcolor{black}{
Those results demonstrate that both two components are important for improving the accuracy in USFDA tasks, which is consistent with the derived theoretical result in the Theorem \ref{the2}. Moreover, the results further reveal that implicit alignment and model smoothness can benefit each other in solving USFDA. }
\begin{figure}[t]
  \centering
  \includegraphics[width=0.5\textwidth]{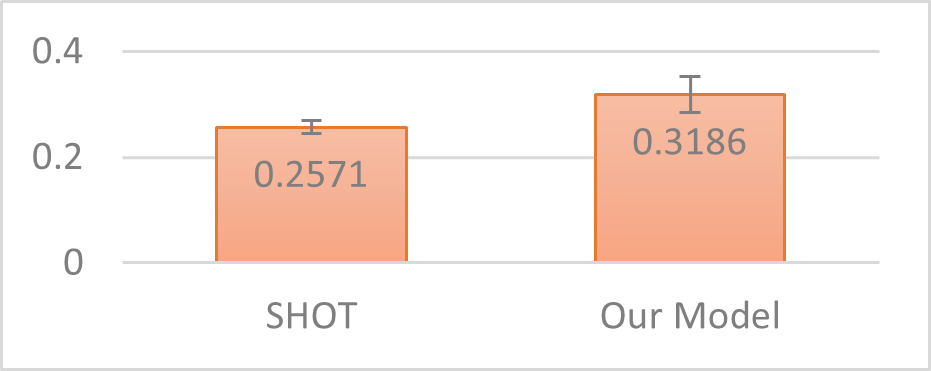}
    \caption{The time cost(s) of SHOT and our Model on the VisDA-C dataset. The error bar represents the standard deviation.}
  \label{figt}
\end{figure}
\subsection{Time Complexity}
\textcolor{black}{
We validate the time complexity of the proposed JN regularizer through the empirical analysis. Specifically, we compare the time cost of our model with SHOT on the large-scale dataset, i.e., \emph{VisDA-C}. The environment is Nvidia RTX 2080Ti with 11G memory. The results are given in Figure. \ref{figt}. Specifically, for each batch, the proposed JN regularization term only introduces additional 0.06s cost on \emph{VisDA-C}. As we can observe, despite its superiority in the performance gain, the time cost paid by the proposed JN regularization term is almost negligible.}
\section{Conclusion}
In this paper, we develop a JN regularizer as a plug-in unit to boost the performance of USFDA. With a few lines of codes, the proposed JN regularizer can significantly improve the performance of the existing USFDAs. It is worth noting that the JN regularization term does NOT need access to the source model and thus can be applied on more challenging black-box USFDA, which will be further studied in our future work.
\label{sec6}

\bibliographystyle{ieeetr}
\bibliography{mybibfile}

%\newpage

 %needed around the contents of the optional argument to biography to prevent


%\bf{If you include a photo:}\vspace{-33pt}
%\begin{IEEEbiography}[{\includegraphics[width=1in,height=1.25in,clip,keepaspectratio]{fig1}}]{Michael Shell}

%\end{IEEEbiography}



%\vfill

\end{document}